\documentclass[conference]{IEEEtran}

\ifCLASSINFOpdf
  \usepackage[pdftex]{graphicx}
  \graphicspath{{Pics/}}
  \DeclareGraphicsExtensions{.pdf,.jpeg,.png}
\else
  \usepackage[dvips]{graphicx}
  \graphicspath{{Pics/}}
  \DeclareGraphicsExtensions{.eps}
\fi
\usepackage{amsmath}
\usepackage{algorithmic}
\usepackage{hyperref}
\newcommand{\vc}[1]{\boldsymbol{\mathbf{#1}}}
\DeclareMathOperator*{\argmin}{argmin}
\DeclareMathOperator*{\diag}{diag}
\DeclareMathOperator*{\oalpha}{\overline{\alpha}}
\usepackage{xcolor,colortbl}
\definecolor{light-gray}{gray}{0.7}
\newcommand{\lightbox}{\color{light-gray}\rule{0.23cm}{0.09cm}}
\usepackage{algorithm}
\usepackage{multirow}

\begin{document}
%
\title{Learning Leading Indicators \\ for Time Series Predictions}



%
\author{\IEEEauthorblockN{Magda Gregorova\IEEEauthorrefmark{1}\IEEEauthorrefmark{2},
Alexandros Kalousis\IEEEauthorrefmark{1}\IEEEauthorrefmark{2}, and
St\'ephane Marchand-Maillet\IEEEauthorrefmark{2}}
\IEEEauthorblockA{\IEEEauthorrefmark{1}University of Applied Sciences and Arts of Western Switzerland, Geneva, Switzerland}
\IEEEauthorblockA{\IEEEauthorrefmark{2}University of Geneva, Switzerland}}


\maketitle

\begin{abstract}
We consider the problem of learning models for forecasting multiple time-series systems together with discovering the leading indicators that serve as good predictors for the system.
We model the systems by linear vector autoregressive models (VAR) and link the discovery of leading indicators to inferring sparse graphs of Granger-causality.
We propose new problem formulations and develop two new methods to learn such models, gradually increasing the complexity of assumptions and approaches.
While the first method assumes common structures across the whole system, our second method uncovers model clusters based on the Granger-causality and leading indicators together with learning the model parameters.
We study the performance of our methods on a comprehensive set of experiments and confirm their efficacy and their advantages over state-of-the-art sparse VAR and graphical Granger learning methods.
\end{abstract}


%
\IEEEpeerreviewmaketitle

\section{Introduction}
\label{sec:intro}

In many application areas multiple indicators are monitored over time.
For planning purposes reliable forecasts of their future developments are needed.
While the monitored series are usually closely related, often there is little understanding of these relationships coming from the specific domain knowledge.

We consider the problem of forecasting such systems of multiple time series from their past evolution together with discovering the main leading indicators within the system that help predicting the future values of most of the series. 
This twofold objective is driven by realistic considerations of the forecasting practice.

In forecasting exercises practitioners typically endeavour to gather as much helpful data as possible.
Many a time this may lead to cluttering their models with indicators with little predictive benefit.
Discovering the leading indicators helpful for forecasting the whole system or its major parts may aid the analysts in understanding the relationships between the series and in simplifying their models and their forecasting systems accordingly (data acquisition, storage, etc.).

While in reality the set of time series included in the system is certainly not exhaustive, 
we deliberately concentrate the analysis on the system series setting aside any possible (and likely) external effects.
Developing our previous line of reasoning, we take it that external confounders cannot be used as predictors for the system either because their identity is unknown (a hidden confounder) and/or because their data are not available.
On the other hand, some of the series included in the system may serve as surrogates for such unavailable data and understanding their contribution to the forecasting model is of utmost interest.

By a leading indicator we understand a time series whose past values are useful for predicting the future of other time series within the system, a notion known in the time-series community as Granger causality (G-causality).
The G-causality structure of a time-series system can be described by a directed graph where each node is a time-series and an edge is a directed G-causal link.

We assume little prior knowledge about the relationships between the series, except that the domain knowledge supports our main assumptions about the existence of lagged dependencies between the series and the general over-specification of the system with only a few main leading indicators. 
In result, we expect the G-causal graph to have rather specific structure: sparsely connected graph with only a few nodes with out-edges leading to most of the other nodes in the graph.
These centrally located nodes are the leading indicators of the system that we wish to discover.

In this paper, we focus on linear vector autoregressive models (VARs) that are simple yet robust models well established in the time-series forecasting literature, e.g. \cite{Lutkepohl2005}, in particular for modelling stationary time series that we concentrate on in this work.
The G-causality is naturally linked to the VAR models through 
the model parameters matrix that can be interpreted as the adjacency matrix of the G-causal graph.
We therefore formulate our twofold problem of forecasting the system together with  discovering the leading indicators as learning VAR models with parameters structure corresponding to our G-causal graph assumptions.

Several methods have been recently proposed for learning the G-causality in the VARs, e.g. \cite{Lozano2009,Shojaie2010,Songsiri2013} (section \ref{sec:relatedwork} has more detailed discussion of the related work).
However, these methods neglect the possibility of shared structures in the lagged dependencies captured by the G-causality graph.

The goal of this paper is to demonstrate that leveraging such shared G-causality structures can improve the VAR learning and yield better forecasting performance of the models, in particular, in the small-sample-large-dimensionality settings which are very common in real multivariate time-series forecasting problems.

We build on the regularized multi-task learning paradigms, e.g. \cite{Evgeniou2004}, and the techniques of structured regularization, e.g. \cite{Bach2012}, to develop two new methods for learning the G-causality in VARs with leading indicators (section \ref{sec:newmethods}).
The first, SingleCluster-VAR (SCVAR), somewhat naively assumes that the leading indicators are useful for forecasting the whole system, that is all the series within the system.
The second, MultiCluster-VAR (MCVAR), assumes more realistically that there are different leading indicators useful for predicting different parts of the system, that is groups of the series. 
These groups are, however, a priori not known and need to be learned together with the model parameters.

To the best of our knowledge these are the first G-causality learning methods that consider common sparse structures in the lagged dependencies.
Also, our approach to learning clusters around their leading indicators together with discovering the graph of G-causality is novel.

To document the validity and the power of our approach we test our methods on a set of artificial and real data experiments and compare their performance with the state-of-the-art methods for learning G-causality in VARs (section \ref{sec:experiments}).
In the synthetic datasets coherent with our structural assumptions our methods clearly out-perform the state-of-the-art competitors.
At the same time, they are robust to violations of such assumptions having predictive performance at least as good as other tested methods in such synthetic data.
Our methods also show superior performance on several real datasets.

\section{Preliminaries}
\label{sec:preliminaries}

\subsection{Vector autoregressive model}
\label{sec:var}

For a set of $K$ time series observed at $T$ synchronous equidistant time points, we write the VAR in the form of a multi-output regression problem as
\begin{equation}\label{eq:VAR}
\vc{Y} = \vc{X W} + \vc{E},
\end{equation}
where $\vc{Y}$ is the $T \times K$ output matrix for $T$ instances and $K$ time series as individual forecasting tasks, $\vc{X}$ is the $T \times Kp$ input matrix so that each row $\vc{x}_{t,.}$ of the matrix is a $Kp$ long vector with $p$ lagged values of the K time series as inputs 
 $\vc{x}_{t,.} = (y_{t-1,1},y_{t-2,1},\ldots,y_{t-p,1},y_{t-1,2},\ldots,y_{t-p,K})'$\footnote{By convention, all vectors in this paper are column vectors.},
and $\vc{W}$ is the corresponding $Kp \times K$ parameters matrix where each column $\vc{w}_{.,k}$ is a model for a single time series forecasting task.
We follow the standard time series assumptions: the $T \times K$ error matrix $\vc{E}$ is a random Gaussian noise matrix with i.i.d. rows $\vc{e}_{t,.} \sim N(\vc{0},\vc{\Sigma})$ and diagonal covariance $\vc{\Sigma}$; the time series are second order stationary and centred (so that we can omit the intercept).

In principle, we can estimate the model parameters by minimising the standard squared error loss
\begin{equation}\label{eq:SquarredLoss}
\mathit{L}(\vc{W}):= \sum_{t=1}^T \sum_{k=1}^K (y_{t,k} - \langle \vc{w}_{.,k},\vc{x}_{t,.} \rangle )^2
\end{equation} 
which corresponds to maximising the likelihood with i.i.d. Gaussian errors.
However, since the dimensionality $Kp$ of the regression problem quickly grows with the number of series $K$ (by a multiple of $p$), usually already relatively small VARs are highly over-parametrised ($Kp \gg T$) and some form of regularisation needs to be used to condition the model learning. 

\subsection{Granger-causality graphs}
\label{sec:granger}

In \cite{Granger1969} the author defines causality based on the predictability of the series: we say that $Z$ \emph{G-causes} $Y$ if we can forecast $Y$ better using the history of $Z$ than without it. 
This concept can be extended to sets of series so that a set of series $\{Z_1, \ldots, Z_l\}$ is said to G-cause series $Y$ if $Y$ can be better predicted using the past values of the set.

The G-causal relationships can be described by a directed graph $\mathcal{G} = \{\vc{\mathcal{V}},\vc{\mathcal{E}}\}$ in which nodes represent the time series in the system, and a directed edge $e_{l,k}$ from $v_l$ to $v_k$ means that time series $l$ G-causes time series $k$, e.g. \cite{Eichler2012}.

In VARs the G-causal relationships are captured within the $\vc{W}$ parameters matrix of model (\ref{eq:VAR}). 
When any of the parameters of the $k$-th task ($k$-th column of the $\vc{W}$) referring to the $p$ past values of the $l$-th input series is non-zero, we say that the $l$-th series G-causes series $k$, and we denote this in the G-causality graph by a directed edge $e_{l,k}$ from $v_l$ to $v_k$ (see figure \ref{fig:WMatrix}).

\subsection{Multi-task learning}
\label{sec:MultiTask}

In multi-task learning (MTL) we benefit from learning several tasks together (as opposed to learning the tasks in isolation) by allowing the models to share information and learn one from another.
The multi-task learning methods find an appropriate balance between task \emph{specificity} and \emph{similarity} of the models in order to maximize the overall predictive performance. 

\section{Learning VARs with leading indicators}
\label{sec:newmethods}

We develop two new methods for learning VARs with leading indicators.
These methods follow two different structural assumptions for the G-causality graphs: \hypertarget{ass1}{SingleCluster-VAR (SCVAR) assumes that there are leading indicators within the system that help predicting \emph{all} the series in the system} (\hyperlink{ass1}{assumption 1}) and aims at identifying those indicators; \hypertarget{ass2}{MultiCluster-VAR (MC) assumes that there are different leading indicators for different clusters of series} (\hyperlink{ass2}{assumption 2}) and aims at learning both, the leading indicators as well as the unknown clusters.

Both our methods exploit the MTL ideas and intertwine the model learning for the system by imposing structural similarity constraints derived from the assumptions above.
This is in contrast to other state-of-the-art VAR and graphical Granger learning methods which, albeit being usually initially formulated as a single multi-output problem, typically decompose into a set of single-output tasks (due to their simple additive structure) which prevents the models from sharing information at learning time.

Though intertwined by the structural similarity, in both methods the models are allowed to diverge as per the specific time-series predictive task a) in their parameter values and b) \hypertarget{ass3}{by allowing  each series to depend on its own lagged values} (\hyperlink{ass3}{assumption 3}).
In this way we balance the structural similarity with the need for task-specificity.

The rather restrictive structural assumption of SCVAR is obviously somewhat naive and limits the usefulness of the method to quite particular types of time-series systems. We present it here mainly as a stepping stone for the more general and flexible MCVAR which we consider to be the main contribution of this paper.

\subsection{SingleCluster-VAR}
\label{sec:scvar}

\begin{figure}[h!]
\centerline{\input{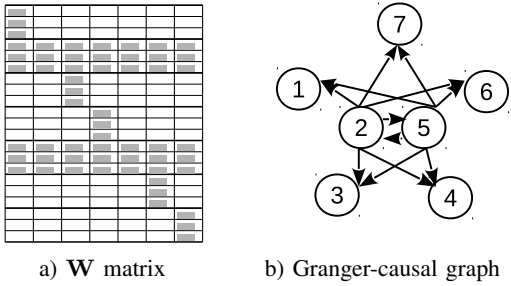}}
\caption{Schematic structure of the SCVAR parameters matrix $\vc{W}$ and the corresponding G-causality graph for a system of $K=7$ series with series 2 and 5 being the leading indicators. In a) the number of lags is $p=3$, the gray cells are the non-zero elements. In b) the self-loops corresponging to the block-diagonal elements in $\vc{W}$ are omitted for clarity of display.}\label{fig:WMatrix}
\vskip -1pt
\end{figure}

We first translate the \hyperlink{ass1}{assumption 1} and \hyperlink{ass3}{3} into the structural properties of the parameters matrix $\vc{W}$ of the VAR model \eqref{eq:VAR}.
According to assumption 1 it shall be a sparse matrix with common non-zero blocks of rows (a block has all lags of a leading indicator) across all the columns (a column is a learning task);
from assumption 3 the matrix shall have full block-diagonal elements (figure \ref{fig:WMatrix}).

Next we bring these structural assumptions into the VAR learning problem.
We partition the input vectors $\vc{x}_{t,.}$ and the parameters $\vc{w}_{.,k}$ in eq. (\ref{eq:SquarredLoss}) into $K$ $p$-long sub-vectors
$\vc{x}_{t,.} = (\tilde{\vc{x}}'_{t,1}, \tilde{\vc{x}}'_{t,2}, \ldots, \tilde{\vc{x}}'_{t,K})'$ 
and 
$\vc{w}_{.,k} = (\tilde{\vc{w}}'_{1,k}, \tilde{\vc{w}}'_{2,k}, \ldots, \tilde{\vc{w}}'_{K,k})'$ so that the $\tilde{\vc{x}}_{t,b}$
contains the $p$ lagged values of series $b$ at time $t$ and $\tilde{\vc{w}}'_{b,k}$ has the parameters for this input sub-vector for the $k$-th predictive task.
We then decompose each of the parameter sub-vectors $\tilde{\vc{w}}_{b,k} = \gamma_{b,k} \tilde{\vc{v}}_{b,k}$ into the product of a non-negative scalar $\gamma_{b,k}$ and a $p$-long vector $\tilde{\vc{v}}_{b,k}$ so that each element $\gamma_{b,k}$ of the 
$K \times K$ matrix $\vc{\Gamma}$ is associated with one $p$-long sub-vector in $\vc{W}$.
Using this decomposition we rewrite the loss function \eqref{eq:SquarredLoss} as
\begin{equation}\label{eq:SquarredLoss2}
\mathit{L}(\vc{\Gamma},\vc{V}) :=  \sum_{t=1}^T \sum_{k=1}^K (y_{t,k} - \sum_{b=1}^K \gamma_{b,k} \langle \tilde{\vc{v}}_{b,k},\tilde{\vc{x}}_{t,b} \rangle )^2,
\end{equation}
where $\vc{V}$ is the $Kp \times K$ matrix with columns $\vc{v}_{.,k} = (\tilde{\vc{v}}'_{1,k}, \tilde{\vc{v}}'_{2,k}, \ldots, \tilde{\vc{v}}'_{K,k})'$ (same shape and decomposition as $\vc{W}$).
It is by controlling the structure in $\vc{\Gamma}$ now that we can control the block-structure in $\vc{W}$ as per our assumptions above.

Since we need to combine two structural assumptions (1 and 3) within a single problem, we further decompose matrix $\vc{\Gamma}$ into two $K \times K$ matrices: 
a sparse matrix $\vc{A}$ with the same sparsity pattern (to be learned) in all its columns ($\vc{\alpha}_{.,k} = \vc{\oalpha}, \forall k$) for assumption 1;
and a fixed diagonal matrix $\vc{B} = \tau \vc{I}$ for assumption 3.
We tie the matrices together by the identity $\vc{\Gamma} = \vc{A} - \diag(\vc{A}) + \vc{B}$, where the middle term frees the diagonal elements from the common learning of $\vc{A}$ and thus leaves them to be specified by $\vc{B}$. 
We set $\tau=1$ in all the following to achieve identifiability\footnote{Note that $\tilde{\vc{w}}_{b,k} = (\tau \gamma_{b,k}) \, (\tilde{\vc{v}}_{b,k}/\tau)$, and $\gamma_{k,k} = \beta_{k,k} = \tau$.}.

We finally formulate the full optimisation problem for SCVAR as a constrained minimisation of loss function \eqref{eq:SquarredLoss2}
\begin{eqnarray}\label{eq:Optimisation1}
& \argmin _{\Gamma,V} \mathit{L}(\vc{\Gamma},\vc{V}), & \\
& \text{ s.t. } \,  \vc{1}'_K \, \vc{\oalpha} = \kappa ;
\, \vc{\oalpha} \geq \vc{0} ; 
\, ||\vc{V}||_F^2 \le \epsilon & \nonumber
\end{eqnarray}
where $||.||_F$ is the Frobenious norm, $\vc{1}_K$ is a $K$-long vector of ones, $\vc{\Gamma} = \vc{A} - \diag(\vc{A}) + \vc{I}$, and $\vc{\alpha}_{.,k} = \vc{\oalpha}$ for all $k$.

In (\ref{eq:Optimisation1}) we impose a standard ridge penalty \cite{Hoerl1970} on the model parameters $\vc{V}$, and a simplex constraint\footnote{$\kappa$ controls the relative weight of each series own past vs the past of all the neighbouring series.} on the common structural vector $\vc{\oalpha}$. It is this simplex constraint which plays the most important role in promoting sparsity in the final $\vc{W}$ matrix at the pre-specified block-level.


In SCVAR we solve the non-convex optimisation problem (\ref{eq:Optimisation1}) to find a local minimum by alternating descent for $\vc{V}$ and $\vc{\oalpha}$ as outlined in algorithm  \ref{alg:SCVAR}. We initialize $\vc{\oalpha}$ evenly so that $\oalpha_b=\kappa/K$ for all $b$. We solve the simplex constrained least squares problem by projected gradient descent method with backtracking stepsize rule \cite{Beck2009}.

\begin{algorithm}[h!]
   \caption{SCVAR \newline {\bfseries Input:} $\vc{Y}, \vc{X}, \lambda, \kappa$; initialize $\vc{\oalpha}$.}
   \label{alg:SCVAR}
\begin{algorithmic}
   \REPEAT
   \STATE Step 1: Solve for $\vc{V}$
   \STATE \hspace{1em} put $\vc{\alpha}_{.,k} = \vc{\oalpha}$ for all columns of $\vc{A}$
   \STATE \hspace{1em} get $\vc{\Gamma} = \vc{A} - diag(\vc{A}) + \vc{I}$
   \STATE \hspace{1em} re-weight input blocks $\vc{z}_{t,b,k} =  \gamma_{b,k} \, \tilde{\vc{x}}_{t,b}$.
	\STATE \hspace{1em} $\argmin_{\vc{v}_{.,k}} ||\vc{y}_{.,k}-\vc{Z}_k \vc{v}_{.,k}||_2^2 + \lambda ||\vc{v}_{.,k}||_2^2 \quad \forall k$
   \STATE Step 2: Solve for $\vc{\oalpha}$
   \STATE \hspace{1em} get block products $h_{t,b,k}=\langle \tilde{\vc{v}}_{b,k}, \tilde{\vc{x}}_{t,b} \rangle$
   \STATE \hspace{1em} get residuals $r_{t,k} = y_{t,k} - h_{t,k,k}$ using own history
   \STATE \hspace{1em} concatenate $h_{t,b,k}$ into $T \times K$ matrix $\vc{H}_k$ and replace $k$th column in $\vc{H}_k$ by zeros
   \STATE \hspace{1em} concatenate $\vc{H}_k$ matrices into $KT \times K$ matrix $\vc{H}$
   \STATE \hspace{1em} $\argmin_{\vc{\oalpha}} ||vec(\vc{R})-\vc{H}\,\vc{\oalpha}||_2^2$, s.t. $\vc{\oalpha}$ on simplex 
  \UNTIL{convergence}
\end{algorithmic}
\end{algorithm}

\subsection{MultiCluster-VAR}
\label{sec:mcvar}

In the MultiCluster-VAR we move from \hyperlink{ass1}{assumption 1} to the more complex (and more realistic) \hyperlink{ass2}{assumption 2} of a cluster structure within the system.

If the cluster structure were known a priori, the models and the leading indicators could be learned by a slight modification of the SCVAR method above: associate each of the $C$ clusters with a structural vector $\vc{\oalpha}^{c}$ and set $\vc{\alpha}_{.,k} = \vc{\oalpha}^c$ for all tasks $k$ belonging to cluser $c$; in step 2 of algorithm \ref{alg:SCVAR} solve for each $\vc{\oalpha}^c$ from the respective $\vc{R}^c$ and $\vc{H}^c$ matrices.

In reality, however, the cluster structure is typically not known a priori and needs to be learned together with the model parameters, which is what our MCVAR method achieves.

\begin{figure}[h!]
\centerline{\includegraphics[width=0.5\columnwidth]{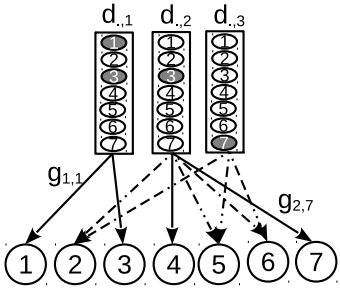}}
\caption{Schema of the role of the $\vc{D}$ and $\vc{G}$ matrices in the low-rank decomposition of matrix $\vc{A}$ in the MCVAR method. Imaginary system with $K=7$ time series and rank $r=3$ matrix $\vc{A}$.
The $\vc{d}$ columns are the sparse cluster prototypes, the non-zero elements are shaded.
The numbered circles in the bottom are the individual task models, the arrows are the elements of the weight matrix $\vc{G}$. Solid arrows have weight 1, dashed arrows have weights between 0 and 1.}\label{fig:DG}
\end{figure}

To bring the \hyperlink{ass2}{assumptions 2} and \hyperlink{ass3}{3} into the VAR learning of MCVAR we use the same structural matrices $\vc{\Gamma} = \vc{A} - \diag(\vc{A}) + \vc{I}$ as in SCVAR.
However, we drop the constraint of strict equality between the columns of $\vc{A}$ (coming from assumption 1) and instead introduce a constraint forcing the columns of $\vc{A}$ to live in a low-dimensional subspace (the clustering assumption 2).
More specifically, we factorize $\vc{A} = \vc{DG}$ into two lower-dimensional matrices $r \leq K$: the $K \times r$ dictionary matrix $\vc{D}$ with the dictionary atoms (columns of $\vc{D}$) representing the cluster prototypes of the dependency structure, and the $r \times K$ matrix $\vc{G}$ with the per-model dictionary weights.
We further impose sparsity promoting simplex constraints on both the dictionary atoms in $\vc{D}$ and the weights of the atoms combination for each model in the columns of $\vc{G}$.
Figure \ref{fig:DG} illustrates the roles of the $\vc{D}$ and $\vc{G}$ matrices in the low-rank decomposition of $\vc{A}$.

Using these factorization matrices, we formulate the MCVAR optimisation problem
\begin{eqnarray}\label{eq:Optimisation2}
& \argmin _{\Gamma,V} \mathit{L}(\vc{\Gamma},\vc{V}), & \nonumber \\
& \text{ s.t. } \vc{1}'_K \, \vc{d}_{.,j} = \kappa \ \forall j ; \; \vc{d}_{.,j} \geq \vc{0}  \ \forall j ; & \\
& \; \vc{1}'_r \, \vc{g}_{.,k} = 1 \ \forall k ;
\; \vc{g}_{.,k} \geq \vc{0}  \ \forall k ;
\; ||\vc{V}||_F^2 \le \epsilon & \nonumber
\end{eqnarray}
where $\vc{A} = \vc{D G}$, and $\vc{\Gamma} = \vc{A} - \diag(\vc{A}) + \vc{I}$.

If in the above the rank is set equal to the size of the system ($r=K$) the model learning disentangles into $K$ independent learning tasks. If $r=1$ the MCVAR is equivalent to our previous SCVAR. In this sense, MCVAR is a more versatile generalization of our two methods which can cater for all of the \hyperlink{ass1}{assumptions 1-3} by setting the value for $r$ correspondingly.

We find a local minimum of the non-convex problem (\ref{eq:Optimisation2}) similarly as in SCVAR by alternating descent for $\vc{V}, \vc{D}$ and $\vc{G}$ as outlined in algorithm \ref{alg:MCVAR}.
We initialise matrices $\vc{D}$ and $\vc{G}$ evenly so that $d_{ij}=\kappa/K$ and $g_{ij}=1/r$ for all $i,j$, 
$\odot$ is the Hadamard product, $\otimes$ is the Kronecker product, and $vec(.)$ is the vectorization operator.

\begin{algorithm}[h!]
   \caption{MCVAR \newline {\bfseries Input:} $\vc{Y}, \vc{X}, \lambda, \kappa$; initialize $\vc{D}, \vc{G}$.}
   \label{alg:MCVAR}
\begin{algorithmic}
   \REPEAT
   \STATE Step 1: Solve for $\vc{V}$
   \STATE \hspace{1em} get $\vc{A} = \vc{DG}$
   \STATE \hspace{1em} get $\vc{\Gamma} = \vc{A} - diag(\vc{A}) + \vc{I}$
   \STATE \hspace{1em} re-weight input blocks $\vc{z}_{t,b,k} =  \gamma_{b,k} \, \tilde{\vc{x}}_{t,b}$.
	\STATE \hspace{1em} $\argmin_{\vc{v}_{.,k}} ||\vc{y}_{.,k}-\vc{Z}_k \vc{v}_{.,k}||_2^2 + \lambda ||\vc{v}_{.,k}||_2^2 \quad \forall k$
   \STATE Step 2a: Solve for $\vc{G}$
   \STATE \hspace{1em} get block products $h_{t,b,k}=\langle \tilde{\vc{v}}_{b,k}, \tilde{\vc{x}}_{t,b} \rangle$
   \STATE \hspace{1em} get residuals $r_{t,k} = y_{t,k} - h_{t,k,k}$ using own history
   \STATE \hspace{1em} concatenate $h_{t,b,k}$ into $T \times K$ matrix $\vc{H}_k$ and replace $k$th column in $\vc{H}_k$ by zeros
   \STATE \hspace{1em} $\argmin_{\vc{g}_{.,k}} ||\vc{r}_{.,k} - \vc{H}_k \vc{D} \, \vc{g}_{.,k}||_2^2$, s.t. simplex $\forall k$
   \STATE Step 2b: Solve for $\vc{D}$
   \STATE \hspace{1em} concatenate $\vc{H}_k$ matrices into $KT \times K$ matrix $\vc{H}$
   \STATE \hspace{1em} get $\hat{\vc{G}} = \vc{G}' \otimes \vc{1}_T \vc{1}'_K$, $\hat{\vc{H}} = \vc{1}'_r \otimes \vc{H}$
   \STATE \hspace{1em} $\argmin_{vec(\vc{D})} ||vec(\vc{R}) - \hat{\vc{G}} \odot \hat{\vc{H}} \, vec(\vc{D})||_2^2$, s.t. simplex on columns of $\vc{D}$
  \UNTIL{convergence}
\end{algorithmic}
\end{algorithm}

To understand better the effects of our methods on the VAR learning we can link our formulations to some other standard learning problems though bearing in mind that these have different aims and different assumptions than ours.
We can rewrite the inner product in the loss function (\ref{eq:SquarredLoss2}) as $\langle \tilde{\vc{v}}_{b,k},\gamma_{b,k} \tilde{\vc{x}}_{t,b} \rangle$. In this ''feature learning`` formulation, the vectors $\vc{\gamma}_{.,k}$ act as weights for the original inputs and, hence, generate new (task-specific) features $\vc{z}_{t,b,k} =  \gamma_{b,k} \, \tilde{\vc{x}}_{t,b}$ (see also algorithms \ref{alg:SCVAR} and \ref{alg:MCVAR}).
Alternatively, we can express the ridge penalty on $\vc{V}$ used in eq. ({\ref{eq:Optimisation2}}) as
$||\vc{V}||_F^2 = \sum_{b,k} ||\tilde{\vc{v}}_{b,k}||_2^2 = \sum_{b,k} 1/\gamma_{b,k}^2 ||\tilde{\vc{w}}_{b,k}||_2^2$. In this ''adaptive ridge`` formulation the elements of $\vc{\Gamma}$, which in our methods we learn, act as weights for the $\ell_2$ regularization of $\vc{W}$.
Equivalently, we can see this as the Bayesian maximum-a-posteriori with Guassian priors where the elements of $\vc{\Gamma}$ are the learned priors for the variance of the model parameters or (perhaps more interestingly) the random errors.

\section{Related work}
\label{sec:relatedwork}

Our work falls into the category of methods for learning the G-causality in VARs.
As shows the list of references in the survey working-paper \cite{Liu2012}, this has been a rather active research area over the last several years.

The closest in spirit to ours are the graphical Granger methods, in particular the lasso-based algorithms \cite{Tibshirani1996} for discovering sparse G-causality graphs.
We use the two best-established ones, the lasso-Granger (LG) of \cite{Arnold2007} and the grouped-lasso-Granger (GLG) of \cite{Lozano2009}, as the state-of-the-art competitors in our experiments.
More recent adaptations of these address the specific problems of determining the order of the models and the G-causality simultaneously \cite{Shojaie2010,Ren2013}, the G-causality inference in irregular \cite{Bahadori2012} and subsampled series \cite{Gong2015}, and in systems with instantaneous effects \cite{Peters2013}.
However, neither of the above methods considers or exploits any common structures in the G-causality graphs as we do in our methods.

Common structures in the dependency are assumed in \cite{Jalali2012} and \cite{Geiger2014} though the common interactions are with unobserved variables from outside the system rather then within the system itself. Also, the methods discussed in these have no clustering ability.
\cite{Songsiri2015} considers common structures across several datasets (in panel data setting) instead of within the G-causality graph of a single dataset.
\cite{Huang2012} assume sparse bi-clustering of the nodes (by the in- and out- edges) to learn fully connected sub-graphs in contrast to our shared sparse structures within the G-causality graph.

Our work builds on the standard regularized multi-task learning and structured regularization techniques developed outside the time-series settings.
Similar block-decompositions of the feature and parameter matrices as we use in our methods have been proposed \cite{Argyriou2008, Swirszcz2012} to promote group structures across multiple models although the methods developed therein have no clustering capability.
Various approaches for learning model clusters are discussed in \cite{Bakker2003,Xue2007,Jacob2008,Kang2011,Kumar2012} of which the latest uses similar low-rank decomposition approach as our method.
Nevertheless, neither of these approaches learns sparse models and builds the clustering on similar structural assumptions as our methods do.

\section{Experiments}
\label{sec:experiments}

We have performed a set of experiments on artificial and real-world datasets to document the performance of our methods and their ability to recover the G-causality structures corresponding to our assumptions 1-3.
We measure the forecasting accuracy by the mean squared error of 1-step-ahead forecasts over a set of hold-out points averaged across all the series in the system. 

To evaluate the ability to recover the expected structures we calculate the number of discovered leading indicators and the number of edges in the G-causality graph.
In the synthetic experiments we calculate the accuracy of the inferred G-causality graph as the percentage of true-positive and true-negative edges on all edges in the graph.

In all the experiments we compare the results of our methods generated from single runs (single initialisation) of our algorithms \ref{alg:SCVAR} and \ref{alg:MCVAR} against simple baseline models: constant prediction (Mean), random walk (RW), univariate autoregression (AR); and the state-of-the-art graphical Granger methods for learning sparse VARs: lasso-Granger (LG) and grouped-lasso-Granger (GLG).

We summarise the experiments and their outcomes in the two subsections below. 
Further details on the experimental settings and the numerical results are available in the Appendix of the online version of this paper.

\subsection{Synthetic experiments}
\label{sec:syntheticexperiments}

\begin{figure}[h!]
\centerline{\includegraphics[width=1\columnwidth]{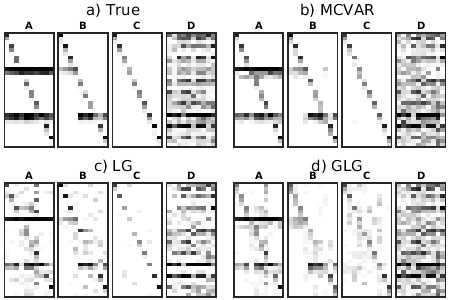}}
\caption{Heatmaps of the VAR parameter matrices $\vc{W}$ for four scenarios of synthetic experiments A-D. a) True parameters used for VAR simulations; b)-d) learned parameters by the respective methods with training samples fixed to 100 instances.}\label{fig:AllTheta4}
\vskip -1pt
\end{figure}

\begin{figure}[h!]
\centerline{\includegraphics[width=1\columnwidth]{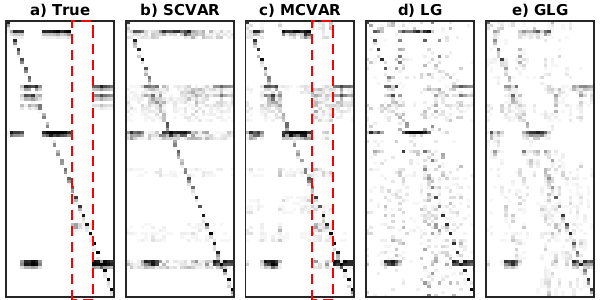}}
\caption{Heatmaps of the VAR parameter matrices $\vc{W}$ for scenario E of the synthetic experiments (see text). a) True parameters used for VAR simulations; b)-e) learned parameters by the respective methods with training samples fixed to 100 instances. The red rectangle emphasizes the 3rd model cluster.}\label{fig:AllTheta1}
\vskip -1pt
\end{figure}

We have created five synthetic datasets (A-E) to explore the behaviour of our methods under different settings.
In the first four (A-D), we examine multiple scenarios of the
G-causality structures as reflected by the parameter matrices $\vc{W}$ depicted in figure \ref{fig:AllTheta4}a) on systems of 10 series ($K=10$).
The first two scenarios correspond to the assumptions of our methods while the other two are on purpose incoherent with them.
Namely, A has the structure of SCVAR; B has the structure of MCVAR with 2 clusters; C is a simple AR model without any dependency structure in between the series; D has a fully connected G-causality graph.
The fifth dataset (E) is to confirm the efficacy of our methods also in larger systems ($K=30$) with more complicated cluster structures (3 clusters in figure \ref{fig:AllTheta1}a)).

All our artificial datasets have been generated from stationary VAR models with $p=3$ and zero centred Gaussian noise with covariance $\vc{\Sigma} = \vc{I}$.
For all the experiments A-E we keep the last 500 data points of the series in the hold-out and train the models using training sets with lengths 30 to 500 data samples (time points). We tune the hyper-parameters by 5-folds inner cross-validation within the training samples\footnote{We use a 10-point logarithmic grid in the interval $[10^{-4},1]$ for $\lambda$ and 4-point logarithmic grid between $[10^{-2},10]$ for $\kappa$.}.

\begin{figure}[h!]
\centerline{\includegraphics[width=1\columnwidth]{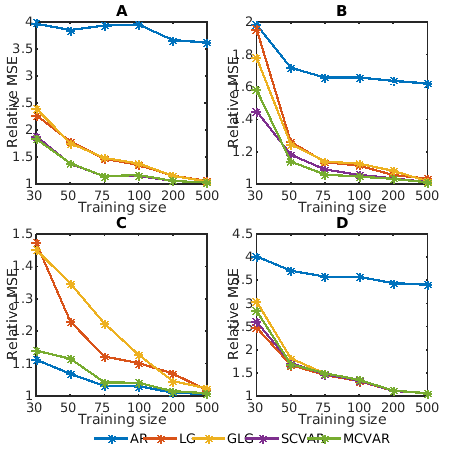}}
\caption{Average 1-step ahead forecast error for A-D scenarios of synthetic experiements measured by relative MSE (see text). For clarity of display Mean and RW methods performing worse by an order of difference are omitted from the plots. In A and C, MCVAR and SCVAR overlap.}\label{fig:mse4}
\vskip -1pt
\end{figure}

\begin{small}
\begin{table}[h!]
\setlength{\tabcolsep}{4pt} 
\caption{Synthetic experiments: relative MSE for 1-step ahead forecasts (hold-out sample average)}
\label{tab:mseSynthetic}
\vskip 0.1in
\begin{center}
\begin{tabular}{c c || c c c c c c }
\hline
\hline
& Size & 30 & 50 & 75 & 100 & 200 & 500 \\
\hline \hline
\multirow{9}{*}{A} &
Mean & 6.81 & 6.81 & 6.81 & 6.81 & 6.81 & 6.81 \\
 & RW & 8.90 & 8.90 & 8.90 & 8.90 & 8.90 & 8.90 \\
 & AR & 3.97 & 3.86 & 3.94 & 3.96 & 3.67 & 3.62 \\
 & LG & 2.28 & 1.79 & 1.47 & 1.36 & 1.16 & 1.06 \\
 & GLG & 2.40 & 1.76 & 1.49 & 1.38 & 1.15 & 1.05 \\
 & SCVAR & 1.89 & 1.38 & 1.15 & 1.15 & 1.06 & 1.03 \\
 & & +++++ & +++++ & +++++ & +++++ & +++++ & +++++ \\
 & MCVAR & 1.84 & 1.39 & 1.14 & 1.17 & 1.06 & 1.03 \\
& & +++++ & +++++ & +++++ & +++++ & +++++ & +++== \\
\hline
\multirow{9}{*}{B} &
Mean & 3.18 & 3.18 & 3.18 & 3.18 & 3.18 & 3.18 \\
& RW & 4.56 & 4.56 & 4.56 & 4.56 & 4.56 & 4.56 \\
& AR & 1.98 & 1.72 & 1.66 & 1.66 & 1.64 & 1.62 \\
& LG & 1.96 & 1.26 & 1.14 & 1.11 & 1.06 & 1.03 \\
& GLG & 1.78 & 1.25 & 1.14 & 1.13 & 1.08 & 1.02 \\
& SCVAR & 1.45 & 1.18 & 1.09 & 1.06 & 1.04 & 1.01 \\
& & +++++ & +++++ & +++++ & +++++ & +++++ & ++++= \\
& MCVAR & 1.58 & 1.14 & 1.06 & 1.05 & 1.03 & 1.01 \\
& & +++++ & +++++ & +++++ & +++++ & +++++ & ++++= \\
\hline
\multirow{9}{*}{C} &
Mean & 1.63 & 1.63 & 1.63 & 1.63 & 1.63 & 1.63\\ 
& RW & 3.55 & 3.55 & 3.55 & 3.55 & 3.55 & 3.55\\ 
& AR & 1.11 & 1.07 & 1.03 & 1.03 & 1.01 & 1.01\\ 
& LG & 1.47 & 1.23 & 1.12 & 1.10 & 1.07 & 1.02\\ 
& GLG & 1.45 & 1.35 & 1.22 & 1.13 & 1.05 & 1.02\\ 
& SCVAR & 1.14 & 1.12 & 1.04 & 1.04 & 1.02 & 1.01\\ 
& & ++=++ & ++=++ & ++=++ & ++=++ & ++=++ & ++=++\\ 
& MCVAR & 1.14 & 1.12 & 1.04 & 1.04 & 1.02 & 1.01\\ 
& & ++=++ & ++=++ & ++=++ & ++=++ & ++=++ & ++=++\\ 
\hline
\multirow{9}{*}{D} &
Mean & 6.56 & 6.56 & 6.56 & 6.56 & 6.56 & 6.56\\ 
& RW & 16.85 & 16.85 & 16.85 & 16.85 & 16.85 & 16.85\\ 
& AR & 4.02 & 3.71 & 3.58 & 3.57 & 3.44 & 3.41\\ 
& LG & 2.48 & 1.67 & 1.45 & 1.32 & 1.11 & 1.06\\ 
& GLG & 3.04 & 1.81 & 1.48 & 1.33 & 1.11 & 1.06\\ 
& SCVAR & 2.60 & 1.71 & 1.47 & 1.33 & 1.11 & 1.06\\ 
& & +++--+ & +++--+ & +++=+ & +++== & +++== & +++==\\ 
& MCVAR & 2.84 & 1.69 & 1.49 & 1.35 & 1.12 & 1.06\\ 
& & +++--+ & +++=+ & +++--= & +++--= & +++== & +++==\\ 
\hline
\multirow{9}{*}{E} &
Mean & 3.73 & 3.73 & 3.73 & 3.73 & 3.73 & 3.73\\ 
& RW & 7.15 & 7.15 & 7.15 & 7.15 & 7.15 & 7.15\\ 
& AR & 1.88 & 1.77 & 1.74 & 1.73 & 1.69 & 1.69\\ 
& LG & 2.41 & 1.81 & 1.79 & 1.51 & 1.19 & 1.08\\ 
& GLG & 2.57 & 2.11 & 1.76 & 1.52 & 1.22 & 1.09\\ 
& SCVAR & 1.88 & 1.56 & 1.49 & 1.25 & 1.11 & 1.05\\ 
& & ++=++ & +++++ & +++++ & +++++ & +++++ & +++++\\ 
& MCVAR & 1.88 & 1.52 & 1.46 & 1.25 & 1.09 & 1.04\\ 
& & ++=++ & +++++ & +++++ & +++++ & +++++ & +++++\\

\hline
\hline
\end{tabular}
\end{center}
Signs below SCVAR and MCVAR methods indicate if the result is significantly better (+), worse (-), or neither (=) as compared to the other methods (in order of the rows in the table) using a one-sided paired-sample t-test at 5\% significance level.
\end{table}
\end{small}

In figure \ref{fig:mse4} and table \ref{tab:mseSynthetic} we compare the predictive performance of the tested methods in terms of mean squared error of 1-step ahead predictions averaged across the 500 hold-out points and the 10 series in the systems A-D.
We use the relative MSE as compared to the forecast error of the VARs with the true parameter matrices used for generating the data $MSE_{relative}=MSE_{method}/MSE_{true}$.
The violet (SCVAR) and green (MCVAR) lines are always below the red (LG) and yellow (GLG) lines in experiments A-C and the difference in the performance is statistically significant at $5\%$ significance level at all the points except 2 (SCVAR and MCVAR vs GLG in experiment B with 500 sample size).
In the D experiments, all the methods perform comparably with our methods being significantly better 5 times, significantly worse 5 times and without a significant difference in the remaining 14 cases (our 2 methods compared vs LG and GLG for 6 sample sizes).
These results confirm the utility of our methods for improving forecasting performance and show that our methods behave well even in systems which violate their initial assumptions.

In all the experiments the relative MSE decreases with the increasing training size and eventually all the methods (except AR in experiments A,B,D) have predictive performance near the true models.
In accordance with our goals, it is in the small-sample settings where the performance of the models differs the most and where the advantage of our methods is the most prominent.

In addition to predictive performance, we monitor the performance of our and the competing methods in terms of our second objective: the recovery of leading indicators and the cluster structure in the G-causality graph. 
Figure \ref{fig:AllTheta4} gives examples of the parameter matrices $\vc{W}$ learned by the tested methods in comparison to the true parameters from which the data of the four synthetic experiments A-D have been generated.
From a visual inspection, the MCVAR seems to be closer to the True than the other two methods, mainly in the A and B experiments.

\begin{figure}[h!]
\centerline{\includegraphics[width=1\columnwidth]{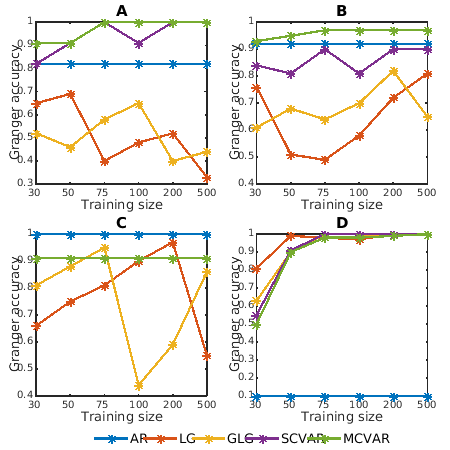}}
\caption{Accuracy of the recovered G-causal graphs for A-D scenarios of synthetic experiements. For clarity of display and consistency with figure \ref{fig:mse4}, Mean and RW methods are omitted from the plots. In C, MCVAR and SCVAR overlap.}\label{fig:acc4}
\vskip -1pt
\end{figure}

We provide a more systematic quantitative evaluation in figure \ref{fig:acc4} where we compare the Granger accuracy of the methods for different training set sizes across the four experiments.
By Granger accuracy we mean the accuracy in correctly identifying the edges in the G-causality graph underlying the learned VAR model (accuracy = (true positive + true negative)/(true positive + true negative + false positive + false negative)).
In the A and B experiments coherent with the initial assumptions of our methods, the violet (SCVAR) and green (MCVAR) lines are always above the red (LG) and yellow (GLG) lines which struggle to recover the correct structures even for larger samples. In the other 2 experiments, the performance of all the methods is comparable.
The simple baseline AR model is included in the graphs for consistency reasons though it does not actually \emph{discover} the structure but has it fixed by construction. 

\begin{figure}[h!]
\centerline{\includegraphics[width=1\columnwidth]{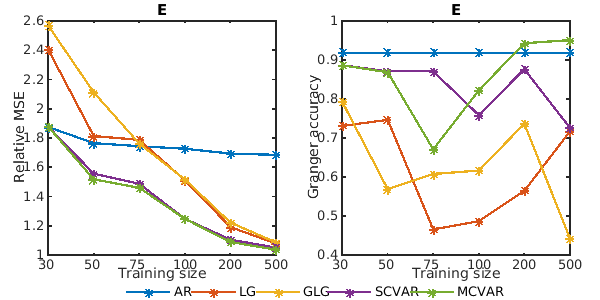}}
\caption{Forecasting performance and the accuracy of the recovered G-causal graphs for scenario E of the synthetic experiments. For clarity of display Mean and RW methods performing worse by an order of difference are omitted from the plots.}\label{fig:mse1}
\vskip -1pt
\end{figure}

For the E dataset, the experimental results are summarised in figure \ref{fig:mse1} using the same metrics as above and 
confirming the superiority of our methods over its competitors in terms of both our objectives, predictive performance (significantly better at $5\%$ sig. level in all 24 cases) as well as the G-causality recovery.

Though the structure of system E is clearly more complicated than what the SCVAR method assumes, we have included it into the evaluation.
Somewhat surprisingly, it performs almost as good as the more complex MCVAR.
This is in part due to the overall strong sparsity of the true models.
The SCVAR therefore suffers only little degradation from including all the leading indicators into all the models due to the shrinking effect of the $\ell_2$ regularization on such miss-specified model parameters.
It is also in part due to the 3rd cluster (emphasized in figure \ref{fig:AllTheta1} in red) which is particularly difficult to discern because of very low parameter values of its leading indicator.

\subsection{Real-data experiments}
\label{sec:realexperiments}

For the real-data experiments we use two sources of data very different in nature in order to verify the performance of our methods in handling more diverse problems.
The first is the database of the Water Services of the US geological survey (\url{http://www.usgs.gov/}) providing (amongst other) the daily averages on water physical discharge along the river streams\footnote{USGS parameter code 00060 - physical discharge in cubic feet per second.}.
The second is a dataset of major US quarterly macro-economic indicators of \cite{Stock2012} (full list in the Appendix of the online version).

\begin{figure}[h!]
\centerline{\includegraphics[width=1\columnwidth]{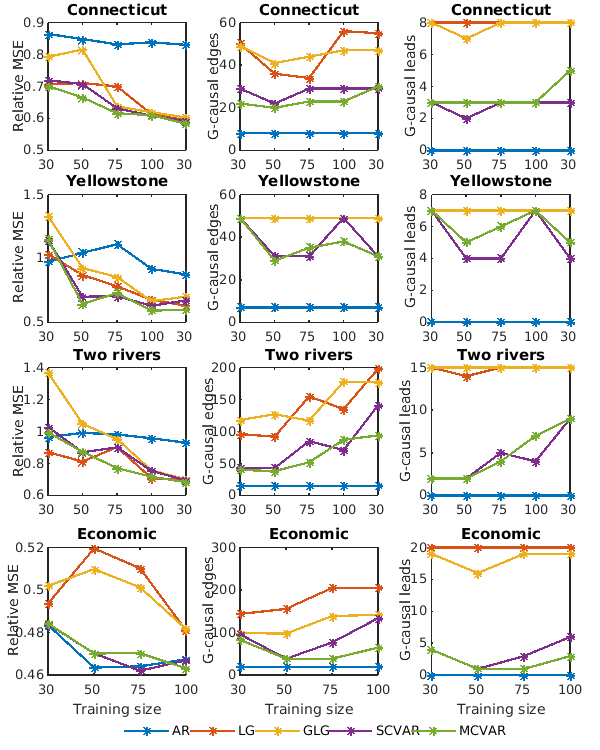}}
\caption{Forecasting accuracy and the sparsity of the G-causality graphs for the 3 river-flow and the Economic data experiments.
The first column: average 1-step ahead forecast error measured by relative MSE; the second column: number of edges in the G-causality graph; the third column: number of leading indicators in the G-causality. For clarity of display the Mean method performing worse by an order of difference in all the experiments is omitted from the plots.}\label{fig:realMse}
\vskip -1pt
\end{figure}

Out of these data-sources, we have constructed four experimental datasets, three from the water data and one from the economic data:
Connecticut with data from $K=8$ measurement sites along the river (650km long in the north-east of the US);
Yellowstone with data from $K=7$ sites (over 1000km long in the west of the US);
Two rivers combining these two into a single system with $K=15$;
Economic with $K=20$ major macro-economic indicators.
We preprocessed all the datasets by standard time-series transformations to achieve stationarity\footnote{Calculating the logs of year-on-year growth for the water data and by differencing or log-differencing for the economic data. Full list of applied transformations is in the Appendix.} and by normalizing.

The forecasting objective for all of these experiments is to predict the values of the series in the next future point, the next day discharge at all of the river sites and the values of the full set of the macro-economic indicators in the next quarter. 
We used a hold-out of the last 100 observations of the series (50 for the Economic data), set $p=5$ ($p=3$ for Economic) and trained the models on the previous 30-300 observations with 5-folds inner cross-validation within the training sets for tuning the hyper-parameters.

\begin{figure}[h!]
\centerline{\includegraphics[width=1\columnwidth]{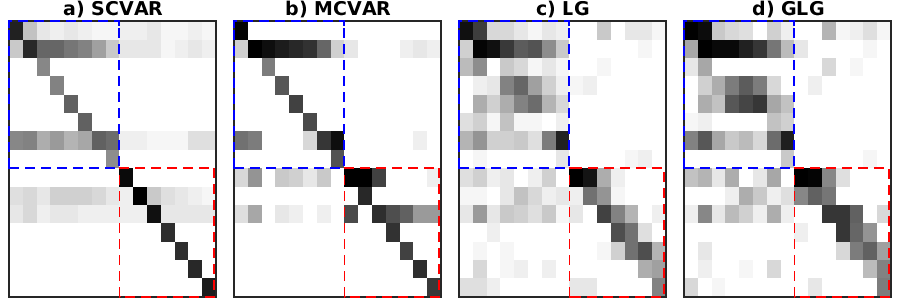}}
\caption{Heatmaps of the learned VAR parameter matrices $\vc{W}$ for the Two rivers experiment with training samples fixed to 100. The measurement sites are ordered in the datasets from the top to the bottom following the rivers from their streams to their mouths. The blue and red rectangles separate the Connecticut and Yellowstone clusters.}\label{fig:ryGranger}
\vskip -1pt
\end{figure}

We provide the summary of the results of the four experiments in figure \ref{fig:realMse}.
The left column shows the relative MSE of the tested methods as compared to the baseline RW model (below 1 means better than RW).
The two other columns summarise the sparsity of the G-causality graphs in terms of the number of edges and the number of discovered leading indicators.
Our methods (violet and green lines) have overall better forecasting performance than the state-of-the-art G-causality learning methods (significantly better at $5\%$ significance 31 times, worse 6 times, no difference 39 times).
At the same time, our methods learn much sparser G-causality graphs uncovering useful structures within the systems as illustrated in figure \ref{fig:ryGranger}. 

\begin{small}
\begin{table}[h!]
\caption{Real-data experiments: relative MSE for 1-step ahead forecasts  (hold-out sample average)}
\label{tab:mseReal}
\vskip 0.1in
\begin{center}
\begin{tabular}{c c || c c c c c }
\hline
\hline
& Size & 30 & 50 & 75 & 100 & 300 \\
\hline \hline
\multirow{8}{*}{\rotatebox[origin=c]{90}{Connecticut}} &
Mean & 2.29 & 2.29 & 2.29 & 2.29 & 2.29\\ 
& AR & 0.86 & 0.85 & 0.83 & 0.84 & 0.83\\ 
& GL & 0.71 & 0.71 & 0.70 & 0.61 & 0.59\\ 
& GLG & 0.79 & 0.82 & 0.64 & 0.62 & 0.60\\ 
& SCVAR & 0.72 & 0.74 & 0.63 & 0.61 & 0.59\\ 
& & ++=+ & ++--+ & +++= & ++== & ++==\\ 
& MCVAR & 0.70 & 0.70 & 0.62 & 0.61 & 0.59\\ 
& & ++=+ & ++=+ & ++++ & ++== & ++==\\ 
\hline
\multirow{8}{*}{\rotatebox[origin=c]{90}{Yellowstone}} &
Mean & 51.91 & 51.91 & 51.91 & 51.91 & 51.91\\ 
& AR & 0.97 & 1.04 & 1.11 & 0.92 & 0.88\\ 
& GL & 1.03 & 0.87 & 0.78 & 0.68 & 0.63\\ 
& GLG & 1.33 & 0.92 & 0.85 & 0.67 & 0.70\\ 
& SCVAR & 1.14 & 0.70 & 0.71 & 0.63 & 0.67\\ 
& & +=== & ++++ & ++=+ & ++== & ++--=\\ 
& MCVAR & 1.14 & 0.64 & 0.73 & 0.59 & 0.60\\ 
& & +=== & ++++ & ++== & ++++ & ++++\\ 
\hline
\multirow{8}{*}{\rotatebox[origin=c]{90}{Two rivers}} &
Mean & 6.93 & 6.93 & 6.93 & 6.93 & 6.93\\ 
& AR & 0.97 & 0.99 & 0.98 & 0.96 & 0.93\\ 
& GL & 0.87 & 0.81 & 0.90 & 0.71 & 0.70\\ 
& GLG & 1.37 & 1.05 & 0.95 & 0.76 & 0.70\\ 
& SCVAR & 1.03 & 0.87 & 0.90 & 0.75 & 0.69\\ 
& & +=--+ & ++--+ & +=== & ++--= & ++==\\ 
& MCVAR & 0.99 & 0.87 & 0.77 & 0.72 & 0.68\\ 
& & +=--+ & ++--+ & ++++ & ++=+ & ++==\\ 
\hline
\hline
\end{tabular}
\vskip 1em
\begin{tabular}{c c || c c c c }
\hline
\hline
& Size & 30 & 50 & 75 & 100\\
\hline \hline
\multirow{8}{*}{\rotatebox[origin=c]{90}{Economic}} &
Mean & 0.60 & 0.60 & 0.60 & 0.60\\ 
& AR & 0.48 & 0.46 & 0.46 & 0.47\\ 
& GL & 0.49 & 0.52 & 0.51 & 0.48\\ 
& GLG & 0.50 & 0.51 & 0.50 & 0.48\\ 
& SCVAR & 0.48 & 0.47 & 0.46 & 0.47\\ 
& & +=== & +=++ & +=++ & +===\\ 
& MCVAR & 0.48 & 0.47 & 0.47 & 0.46\\ 
& & +=== & +=++ & +=++ & +===\\ 
\hline
\hline
\end{tabular}

\end{center}
Signs below SCVAR and MCVAR methods indicate if the result is significantly better (+), worse (-), or neither (=) as compared to the other methods (in order of the rows in the table) using a one-sided paired-sample t-test at 5\% significance level.
\end{table}
\end{small}

\section{Conclusions}
\label{sec:conclusions}

We have developed two new methods for learning sparse VAR models with shared structures in their Granger causality graphs based on the leading indicators of the system, a problem that had not been previously addressed by the G-causality learning methods.

The first method, the somewhat simpler and more naive SCVAR, serves as a building stone for the more complex and versatile MCVAR which achieves several learning objectives simultaneously: good forecasting performance of the models, discovery of clusters within the G-causality graphs and of their leading indicators.

We have confirmed the efficacy of our methods in a series of synthetic and real-data experiments where our methods systematically outperformed the state-of-the-art methods in both the predictive performance and the sparsity of solutions.

Extensions of our methods worth exploring in future work are the analysis of contemporaneous dependencies, relaxation of the stationarity assumption and non-linearities in the dependencies and the time-series generative processes.



\vskip 3em
\bibliographystyle{IEEEtran}
\bibliography{mg_icml2016}

\end{document}